# Potential Penetrative Pass (P3)

Paper Track

Hadi Sotudeh[1]

## Introduction

To score goals in football, a team needs to move forward on the pitch (Michalczyk, 2020) and there are various ways to do so. Depending on the game plan & philosophy; some teams prefer to play long balls from either wings or defense such as Burnley FC. Others, prefer to **penetrate** in depth with passes and outplay the opponent players such as Chelsea FC.

To objectively & in an automated way evaluate how teams play **penetrative passes** compared to the number of times they had the **potential** to do so, I introduce the concept of **Potential Penetrative Pass (P3)** in this study.

## Definitions

I define a P3 as when a player is in a passing moment with his/her foot and there is at least a teammate to receive the ball inside the convex hull created by the opponent players in front of the ball (excluding their goalkeeper). To only consider valuable penetrative passes, I ignore passes in the first 3rd of the defense and the last 4th of the pitch, see Figures 1 and 2.

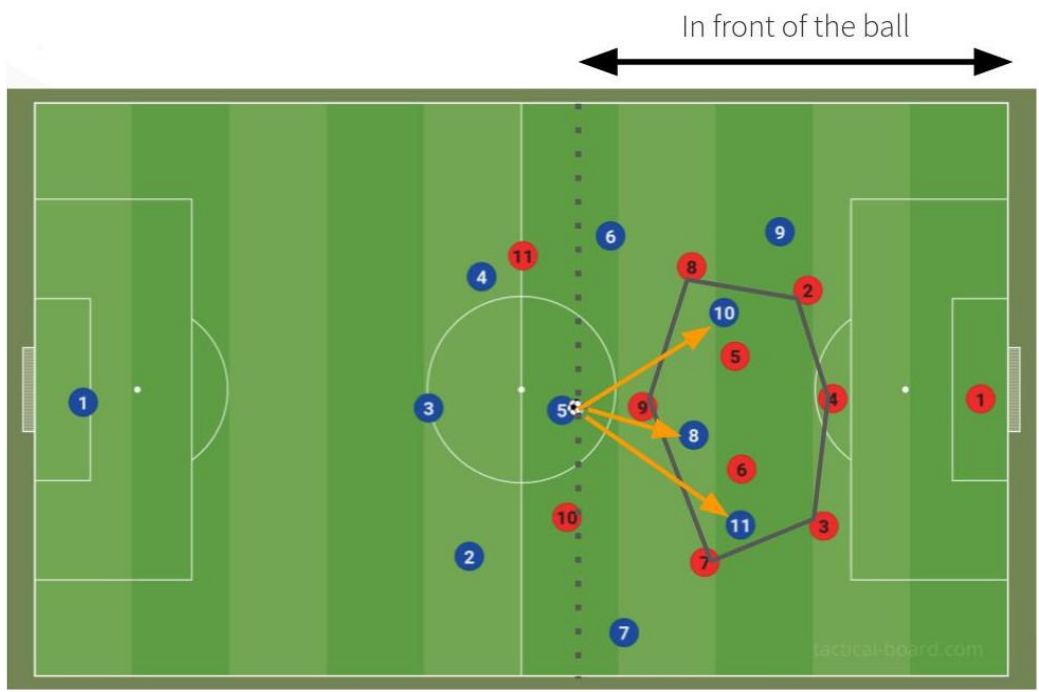

Figure 1: Potential Penetrative Pass Example

---

[1] hadisotudeh1992@gmail.com

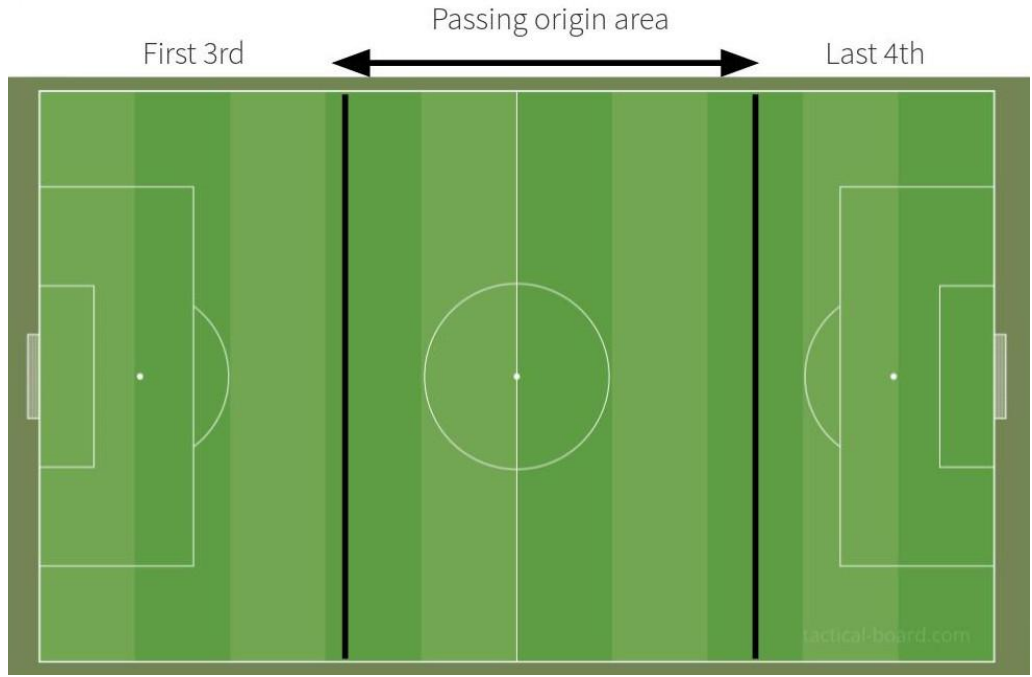

Figure 2: Potential Penetrative Pass Area

## Task

In this study, I would like to specifically answer the following question and introduce the applications it can have:

*Given a P3 moment, what is the probability it becomes a penetrative pass?*

## Data

To conduct this study, I was provided with [StatsBomb 360](#) & [event](#) datasets for Bundesliga and LaLiga in the 2020-2021 season (only the matches of the top 10 teams).

## Approach

Assuming input X, as the information I have in a passing moment, I would like to predict whether the pass becomes penetrative or not (output Y). To perform this task, which is known as binary classification in the machine learning (ML) literature, I explored the following methods & compared their results to select the final solution:
1. ML algorithms trained solely on event data features
2. Convolutional Neural Networks (CNNs) trained on the image representation of the P3 moment, which is created from both 360 & event data

It is worth mentioning that the Bundesliga matches, which contained 245 matches of 360 & event datasets (79,000 P3 moments) were used to train and validate the model (The first 80% of the matches in the training set and the last 20% in the validation set).

Since the dataset is imbalanced (88% of the P3 moments are not penetrative), accuracy can't be a proper evaluation metric to use. Therefore, I employed Area Under the Receiver Operating Characteristic Curve (AUC ROC) because it is insensitive to imbalanced datasets.

## Method 1: ML algorithms trained solely on event data features

As shown in Figure 3, StatsBomb event data has information on the pass location and whether the pass was given under the opponent's pressure.

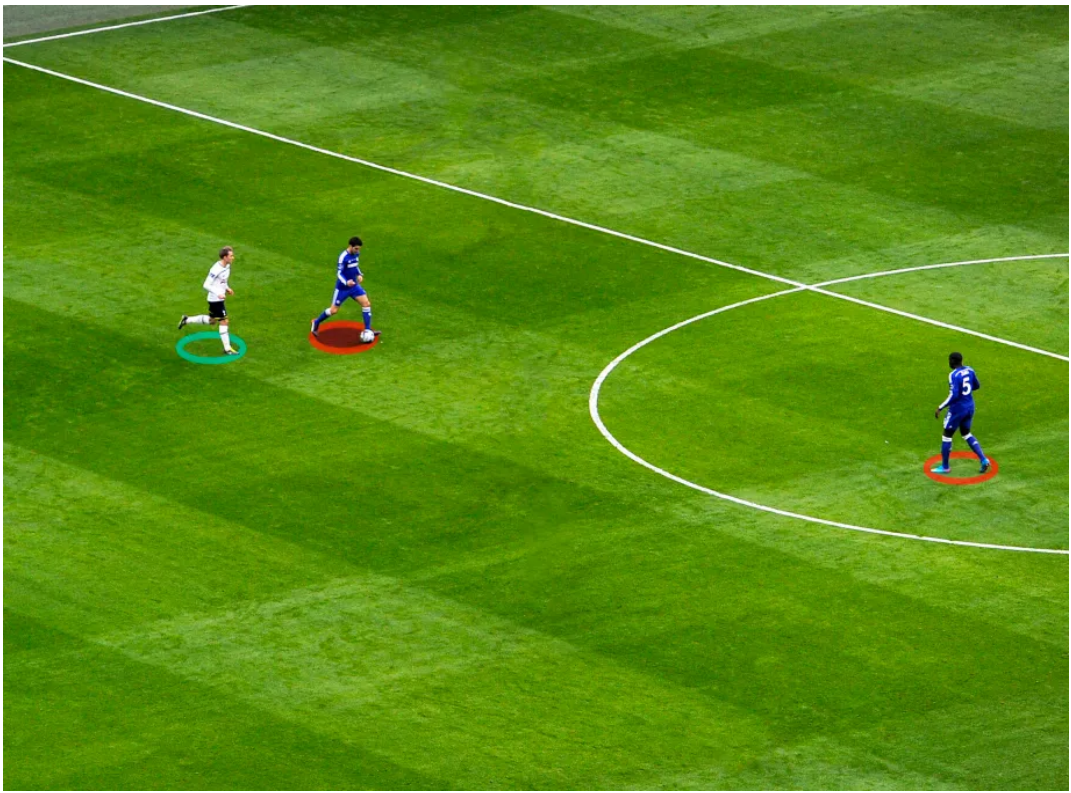

Figure 3: An example of what StatsBomb event data includes

This data also has passer-dependent information such as body type (right or left foot) or where the end location of the pass is, but I don't use these two attributes because one knows their values after the pass is conducted. It is also not scientifically correct to use post-event features to train the models, which is called *data leakage* in the ML literature.

Table 1 shows the results of some well-known ML algorithms trained on the event features after conducting hyper-parameter tuning.

Table 1: Results of the ML algorithms trained solely on event features

| Model | Training (AUC ROC) | Validation (AUC ROC) |
|---|---|---|
| Random Forest | 0.55 | 0.55 |
| XGBoost | 0.54 | 0.55 |
| LGBM | 0.54 | 0.55 |
| CatBoost | 0.55 | 0.55 |
| Extra Tree | 0.56 | 0.55 |

The best AUC ROC is 0.55 (slightly better than the baseline), but it is still a poor performance.

## Method 2: CNN trained on the P3 images

As shown in Figure 4, StatsBomb 360 data adds more context to the event data by providing location information of the players visible on the TV camera.

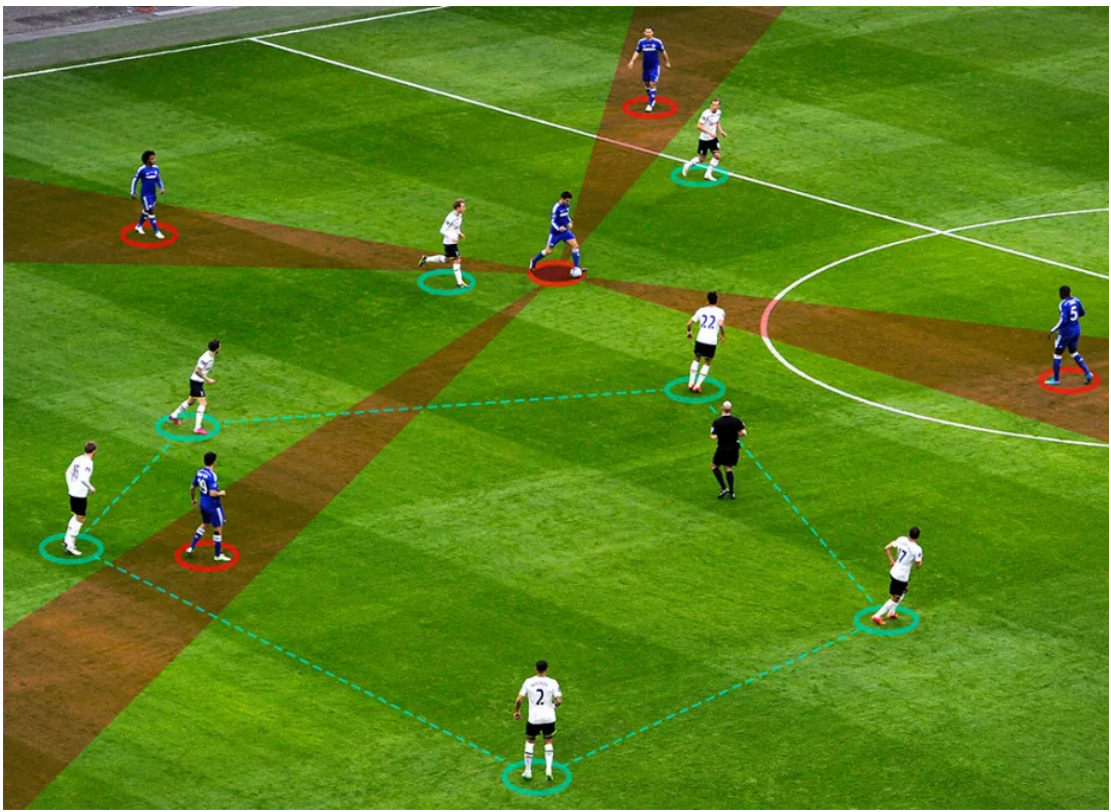

Figure 4: An example of what StatsBomb 360 data includes

This dataset opens up new opportunities for this study. Classifying images into different categories such as cats vs. dogs is a well-known computer vision task addressed by convolutional neural networks (CNNs). Following a similar approach to distinguish between penetrative and non-penetrative passes, I can get probabilities for each category. First, I represent each P3 moment in an image shown in Figure 5.

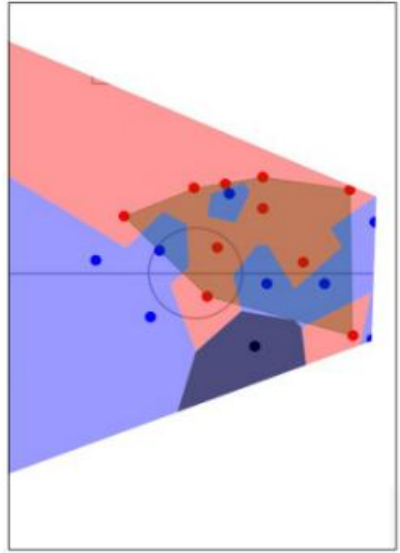

Figure 5: A P3 image representation

As the 360 data has only information of the players visible on the camera angle, the image shown in Figure 5 also shows only that part of the pitch where the team in ball possession is always in **blue** and playing from bottom to top of the pitch and the passer is in **black**. The off-ball team is always depicted in **red** playing from top to bottom of the pitch.

Since classifying an image that only has some **blue**, **red**, and **black** points is a difficult task for a CNN model, I decided to enrich each image with the space each team controls known as "pitch control" models (Spearman, 2016). A simple Voronoi diagram was used to show the controlled space by the **blue** team, the **red** team, and the **passer**. Moreover, I draw the target convex hull in **green** hoping the overlapping colors provide enough information to the classifier.

To not start from scratch and to stand on the shoulders of giants, I made use of a technique called *Transfer Learning* (Bozinovski & Fulgosi, 1976). Following this technique, a pre-trained ResNet model (He et al., 2016) with 34 layers, which was trained on millions of images known as ImageNet (Deng et al., 2009), is fine-tuned for this specific task. Therefore, our initial model is capable of transferring that knowledge such as finding edges and after the fine-tuning process, it will be able to find the specific features required for this task.

After training the model for 8 epochs with a GPU (Nvidia P100) for 2 hours, the training loss becomes 0.30 and the validation loss ends up at 0.33 which suggests that there is no serious overfitting to the training data. The validation AUC ROC is also 0.73, as shown in Figure 6, which is a significant improvement compared to the performance of the previous method.

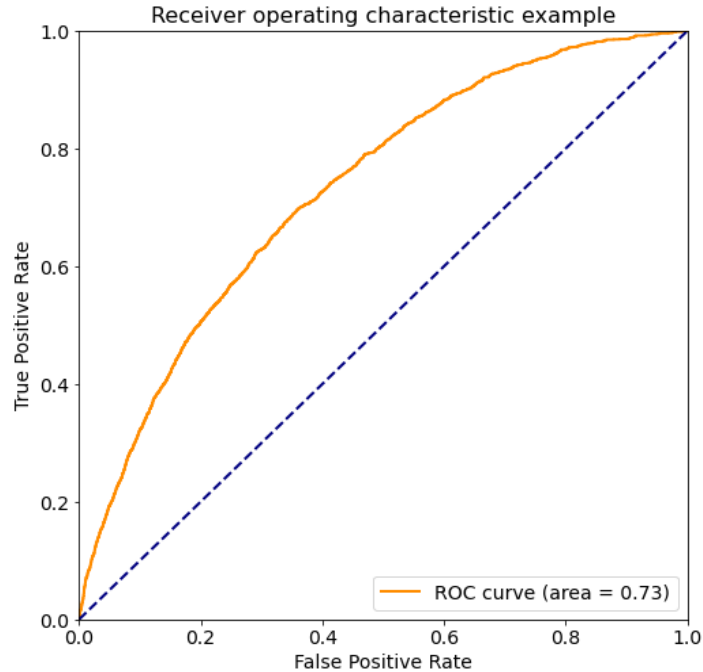

Figure 6: CNN model AUC ROC on the validation set

Since I am also interested in distinguishing between penetrative and non-penetrative passes, selecting a probability threshold is needed. To do so, the corresponding threshold to the closest point on the AUC ROC curve (shown in Figure 6) to point (1,0) in terms of the Euclidean distance is chosen. So, any probability greater than or equal to **10.38%** is classified as a penetrative pass. Figure 7 shows that the model rarely outputs a probability higher than 60% and the median is also around 10%. This is also clear in the quantile-based calibration plot shown in Figure 8

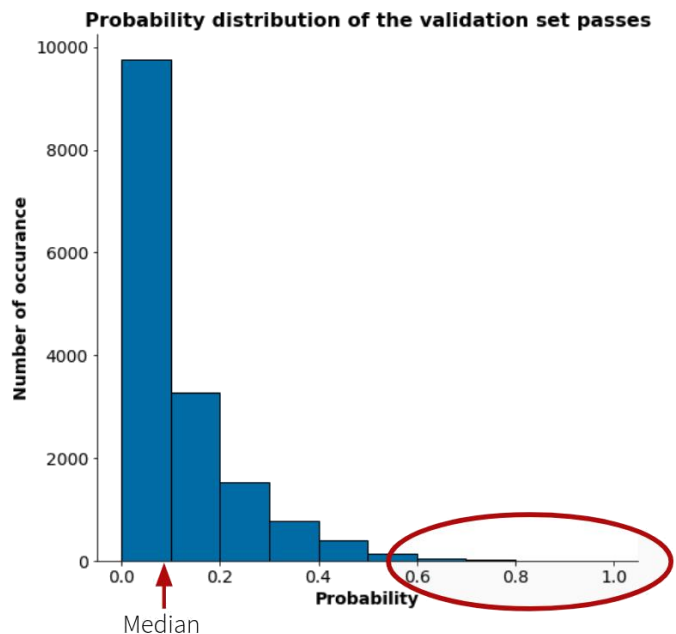

Figure 7: Probability distribution of the model on the validation images

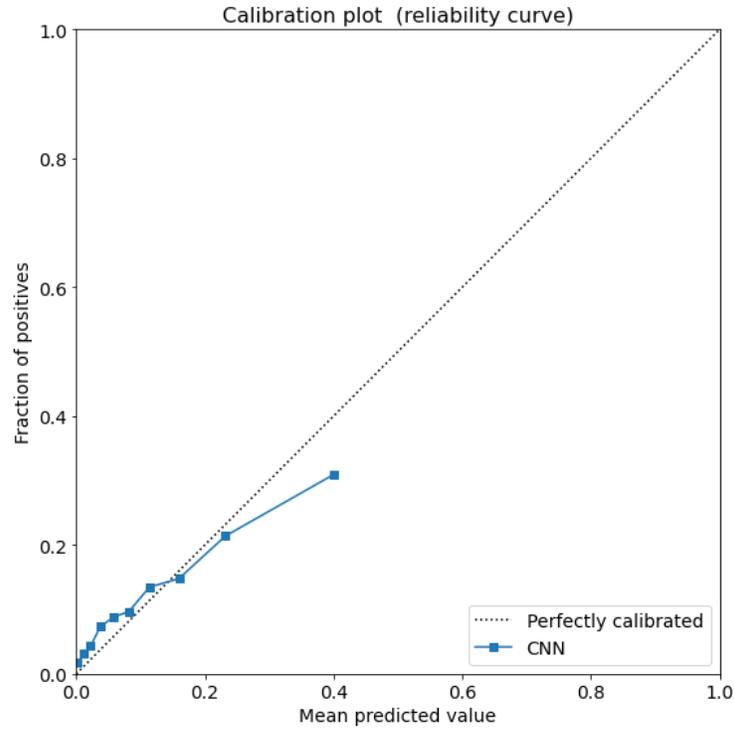

Figure 8: Quantile-based calibration plot of the CNN model

To have a better understanding of how the model classifies images, Figure 9 shows its confusion matrix on the validation set (15,976 images).

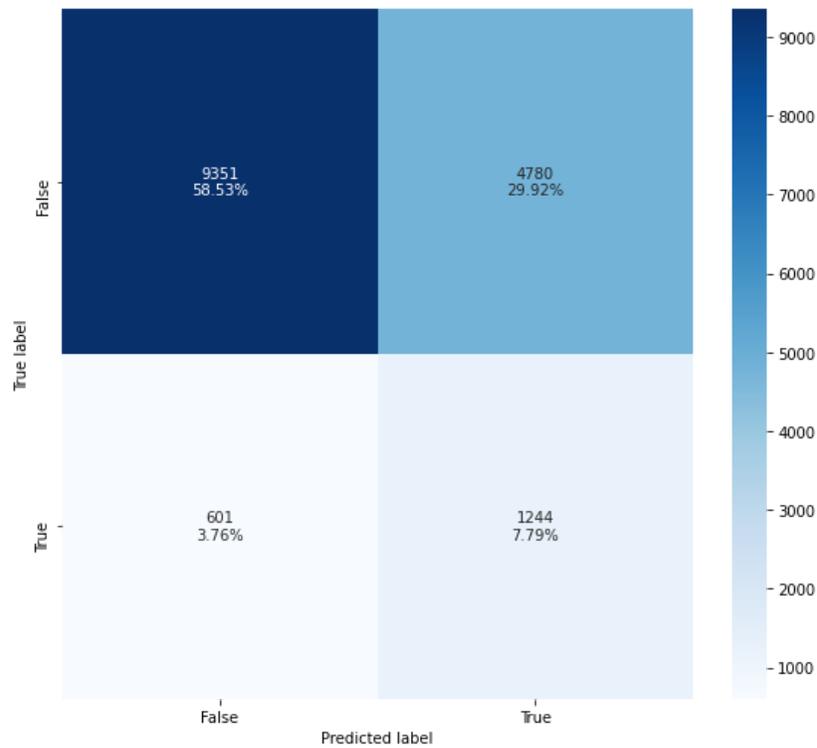

Figure 9: Confusion matrix of the model outputs on the validation set

The model with the chosen probability threshold (10.38%) can correctly detect 9,351 non-penetrative passes and 1,244 penetrative ones. It misclassifies 601 penetrative passes as non, and 4,780 non-penetrative passes as penetrative ones, which can be because some players outperform the average expectation or make other decisions on the pitch.

For example, Figure 10 shows an example of how the model evaluates a P3 moment.

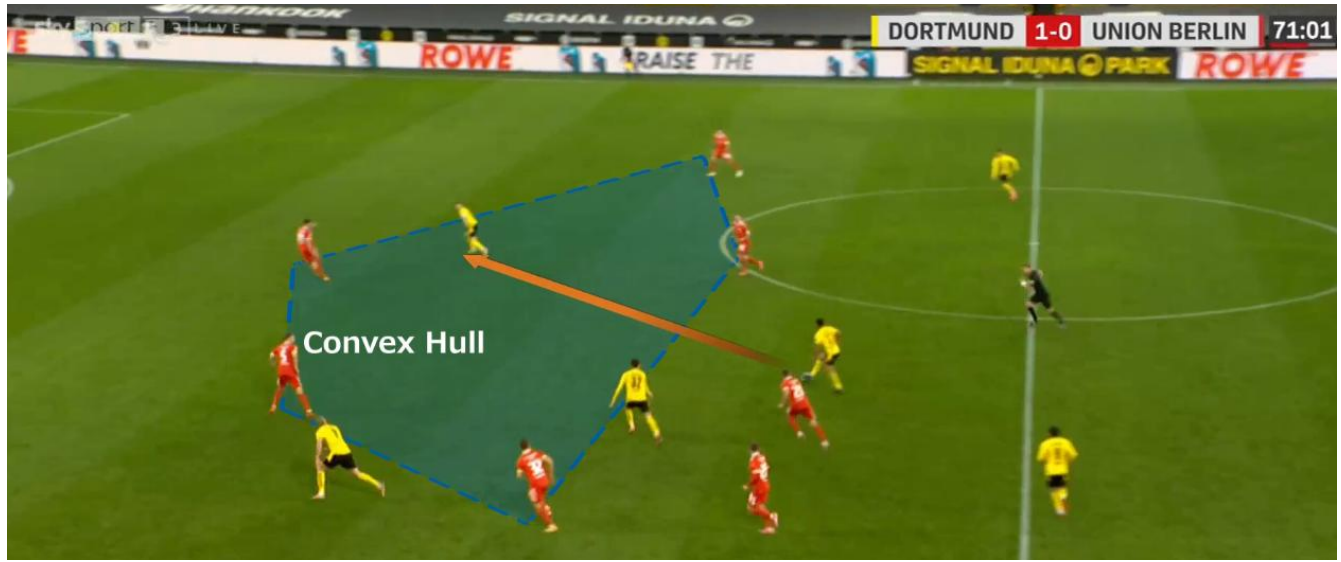

Figure 10: A non-penetrative pass example with a high probability to become a penentrative one

## Applications

I applied the model to the 284 LaLiga matches to highlight the various applications P3 can have:
1. Player recruitment
2. Performance & Opponent Analysis
3. Media

For the recruitment application, I group the players into defenders, midfielders, and under 23. Then, I keep players who had a playtime of at least half of the matches in the season and also had at least the median of the penetrative passes in their group. Instead of only counting the number of penetrative passes to evaluate players, I introduce a new KPI called "P3 Percentage":

$$P3\ percentage = \frac{\#\ successful\ penetrative\ passes}{\#\ potential\ penetrative\ passes}$$

Employing this KPI, I can contextualize penentrative passes by comparing it to the number of times a player could have had a penentrative pass.

Figures 11, 12, and 13 show this KPI calculated for players in each group.

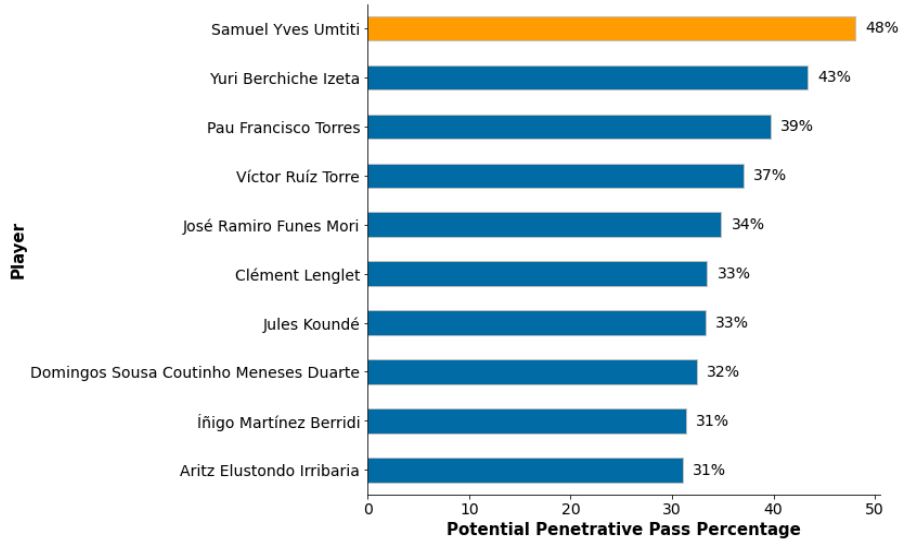

Figure 11: P3 Percentage results of the LaLiga defenders

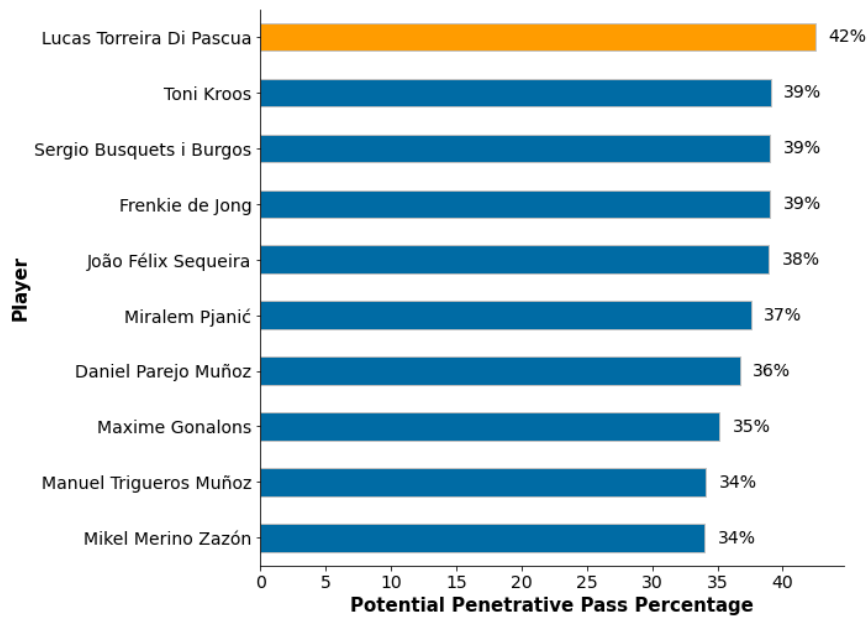

Figure 12: P3 Percentage results of the LaLiga midfielders

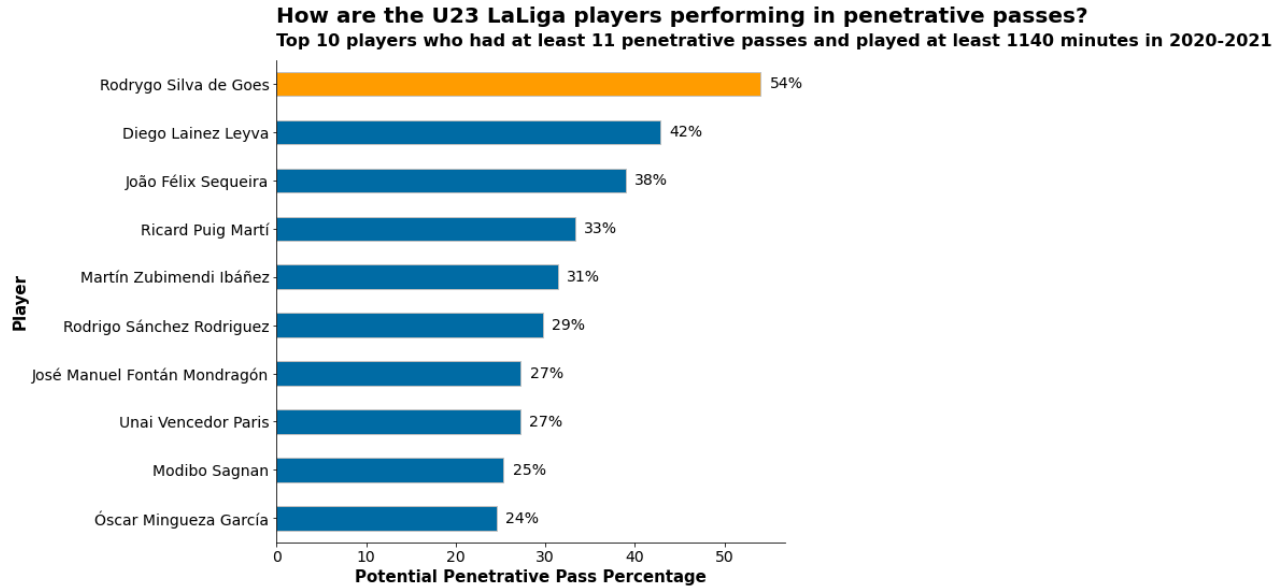

Figure 13: P3 Percentage results of the LaLiga U23 players

I can do the same analysis at the team level for the performance & opponent analysis application. As shown in Figures 14 and 15, Barcelona FC plays a penetrative pass 32% of the time they can and they are on top of the league in that aspect. In addition, Sevilla FC only gives 52 penentrative pass opportunities per match to their opponents and they are the best defending team based on this metric. One should consider the limitations of this KPI such as the camera angle biases and off-camera players that can make the convex hull calculation wrong. These limitations are elaborated in the "discussion" section later.

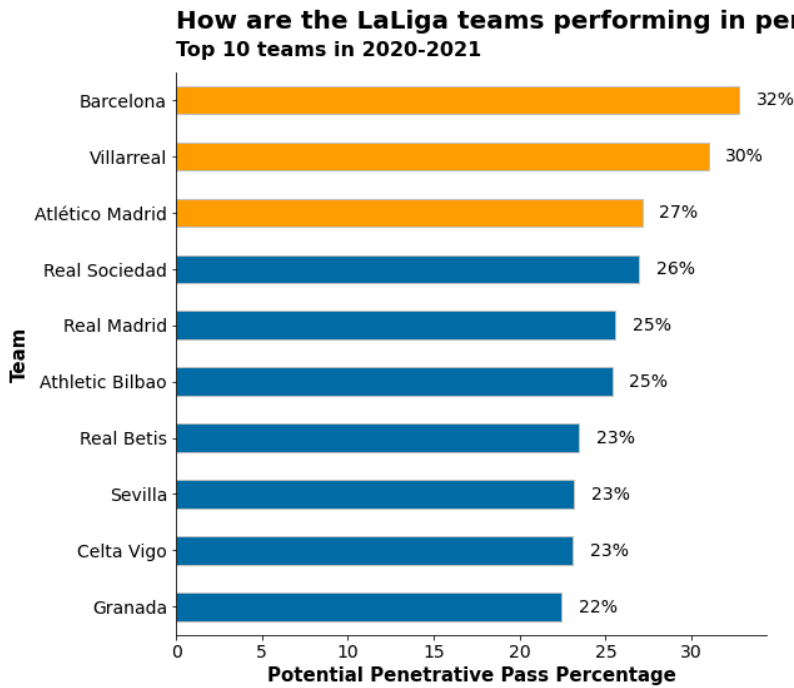

Figure 14: P3 Percentage results for how the top 10 LaLiga teams attack

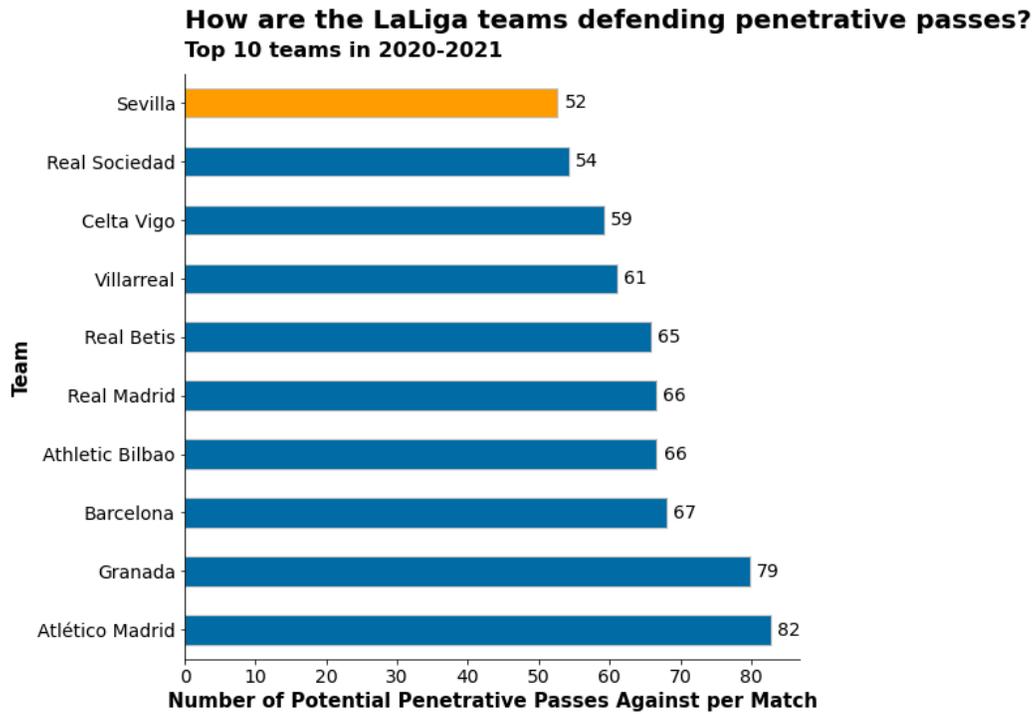

Figure 15: P3 Percentage results for how the top 10 LaLiga teams defend

For the Performance & Opponent Analysis application, one should also consider that the coaching staff spends a lot of time trawling through video footage, which is a time-consuming and tedious task to perform pre & post-match. To support them in the match preparation process, I developed a dashboard called "P3 explorer" where they can analyze individual P3 moments by creating filters, as shown in Figure 16.

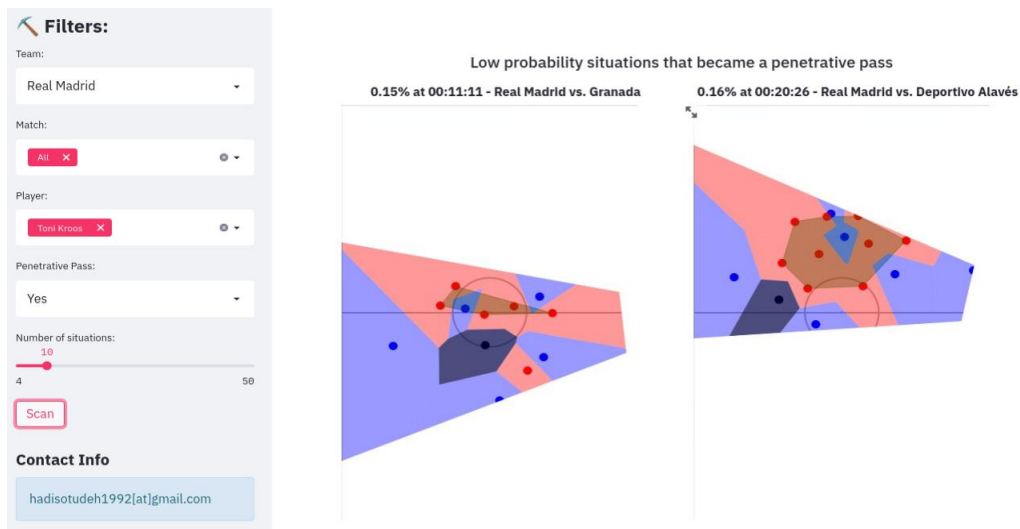

Figure 16: P3 explorer demo

If I also have access to the video footage, "P3 explorer" can be integrated with that source and show clips of the P3 moments instead of the static images, see Figure 17.

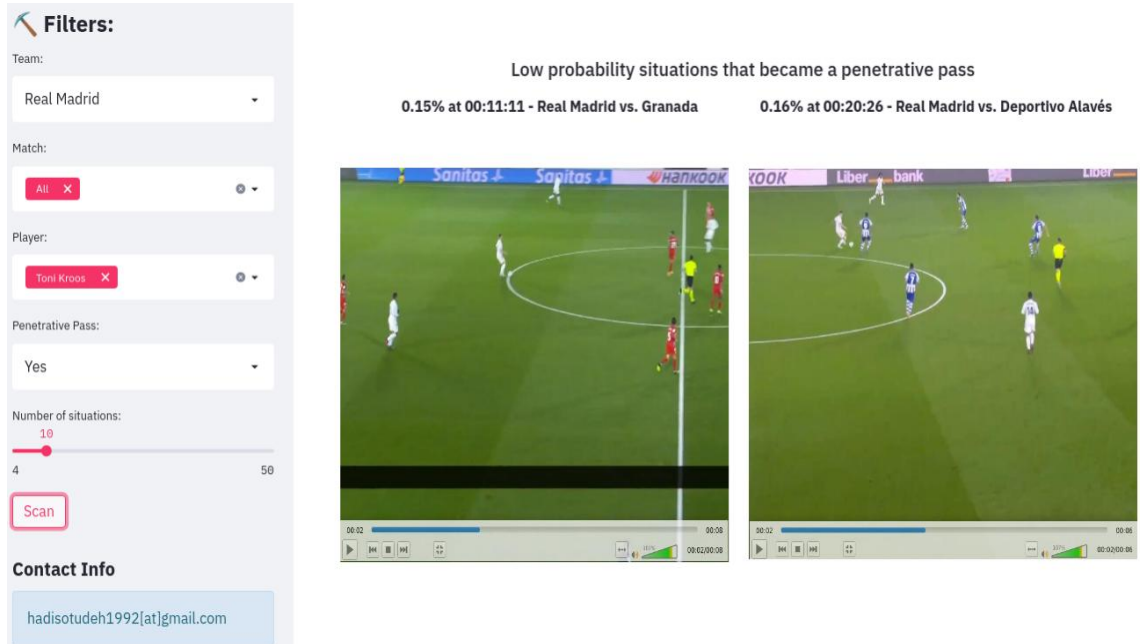

Figure 17: P3 explorer demo integrated with video clips

In this way, it can provide more valuable results for both coaches and media outlets, as video is what humans can consume more easily.

## Discussion & Future Work

The camera angle plays an important role in determining the convex hull, which can hurt the model in situations that only a few off-team players are visible and the ground truth convex hull is different from the visible one on the camera. This is a known issue caused by not having access to the full freeze-frame in a passing moment, but one should keep in mind the 360 data freeze frames usually don't have this issue in passes in the off-ball team half.

The 360 data also does not include information on the body orientation of the players in the passing moment. Therefore, it can be the case that the model says a player has a high potential for a penentrative pass, while his orientation was in the opposite direction to give that pass.

These issues can affect different players/teams to different extents. I hope in the near future, I can reproduce the same study using the full freeze-frame in on-ball moments considering the advances in data collection technologies that potentially can mitigate these problems.

Regarding the possible directions for future work, the following list is prepared:

1. Experiment with different image representations
2. Create a tool to experiment with the impact of changing player positions on the probability output of the P3 model

Finally, the framework explained in this study can be applied to other concepts such as the *potential cross* or *potential counter-attack* assuming the relevant data is provided.

# References

Bozinovski, S., & Fulgosi, A. (1976). The influence of pattern similarity and transfer learning upon the training of a base perceptron B2. *Proceedings of Symposium Informatica*.

Deng, J., Dong, W., Socher, R., Li, L.-J., Li, K., & Fei-Fei, L. (2009). Imagenet: A large-scale hierarchical image database. *IEEE conference on computer vision and pattern recognition*, 248--255.

He, K., Zhang, X., Ren, S., & Sun, J. (2016). Deep Residual Learning for Image Recognition. *IEEE Conference on Computer Vision and Pattern Recognition (CVPR)*, 770-778.

Michalczyk, K. (2020, July 1). *How impactful are line-breaking passes?* OptaPro Forum 2020. Retrieved October 28, 2022, from https://www.statsperform.com/resource/how-impactful-are-line-breaking-passes/

Spearman, W. (2016). *Quantifying Pitch Control*. OptaPro Analytics Forum. 10.13140/RG.2.2.22551.93603